\documentclass[10pt,twocolumn,letterpaper]{article}

\usepackage{iccv}
\usepackage{times}
\usepackage{epsfig}
\usepackage{graphicx}
\usepackage{amsmath}
\usepackage{amssymb}
\usepackage{rotating}

\usepackage[percent]{overpic}

\usepackage[pagebackref=false,breaklinks=true,letterpaper=true,colorlinks,bookmarks=false]{hyperref}

\iccvfinalcopy 

\def\iccvPaperID{2215} 

\newcommand{\MNAME}{Multiscale CNN Regression}  
\newcommand{\MABBR}{MSCR}                       

\begin{document}

\title{%
   \resizebox{\textwidth}{!}{%
      Direct Intrinsics: %
         Learning Albedo-Shading Decomposition by Convolutional Regression%
   }%
}%

\author{
   Takuya Narihira\\
   UC Berkeley / ICSI / Sony Corp.\\
   {\tt\small takuya.narihira@jp.sony.com}\\
   \and
   Michael Maire\\
   TTI Chicago\\
   {\tt\small mmaire@ttic.edu}\\
   \and
   Stella X. Yu\\
   UC Berkeley / ICSI\\
   {\tt \small stellayu@berkeley.edu}
}

\maketitle

\begin{abstract}
We introduce a new approach to intrinsic image decomposition, the task of
decomposing a single image into albedo and shading components.  Our strategy,
which we term \textbf{direct intrinsics}, is to learn a convolutional neural
network (CNN) that directly predicts output albedo and shading channels from
an input RGB image patch.  Direct intrinsics is a departure from classical
techniques for intrinsic image decomposition, which typically rely on
physically-motivated priors and graph-based inference algorithms.

The large-scale synthetic ground-truth of the MPI Sintel dataset
plays a key role in training direct intrinsics.  We demonstrate results on
both the synthetic images of Sintel and the real images of the classic MIT
intrinsic image dataset.  On Sintel, direct
intrinsics, using only RGB input, outperforms all prior work, including methods
that rely on RGB+Depth input.  Direct intrinsics also generalizes across
modalities; it produces quite reasonable decompositions on the real images of
the MIT dataset.  Our results indicate that the marriage of CNNs with synthetic
training data may be a powerful new technique for tackling classic problems in
computer vision.

\end{abstract}

\section{Introduction}
\label{sec:intro}

Algorithms for automatic recovery of physical scene properties from an input
image are of interest for many applications across computer vision and
graphics; examples include material recognition and re-rendering.  The
intrinsic image model assumes that color image $I$ is the point-wise product
of albedo $A$ and shading $S$:
\begin{align}
   I &= A \cdot S
   \label{eq:intrin}
\end{align}
Here, albedo is the physical reflectivity of surfaces in the scene.
Considerable research focuses on automated recovery of $A$ and $S$ given as
input only color image $I$~\cite{horn:retinex74,grosse:intrinsic09}, or given
$I$ and a depth map $D$ for the
scene~\cite{Lee12,barron:intrinsic13,barron:intrinsic15,CK:ICCV:2013}.  Our work falls into the
former category as we predict the decomposition using only color input.  Yet,
we outperform modern approaches that rely on color and depth
input~\cite{Lee12,barron:intrinsic13,barron:intrinsic15,CK:ICCV:2013}.

\begin{figure}
   \begin{center}
      \includegraphics[width=0.95\linewidth, clip=true, trim=0.25in 1.0in 0.25in 0.70in]{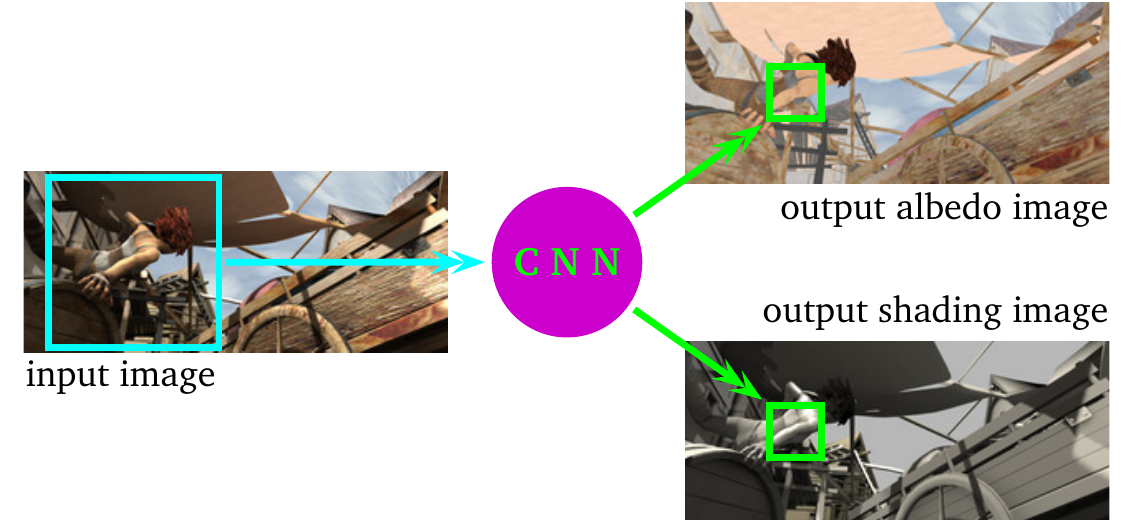}
   \end{center}
   \vspace{-0.02\linewidth}
   \caption{
      \textbf{Direct intrinsics.}
         We construct a convolutional neural network (CNN) that, acting
         across an input image, directly predicts the decomposition into
         albedo and shading images.   It essentially encodes nonlinear
         convolutional kernels for the output patches (green boxes) from
         a much larger receptive field in the input image (cyan box).
         We train the network on computer graphics generated images from
         the MPI Sintel dataset~\cite{Sintel} (Figure~\ref{fig:sintel}).
   }
   \label{fig:overview}
\end{figure}

We achieve such results through a drastic departure from most traditional
approaches to the intrinsic image problem.  Many works attack this
problem by incorporating strong physics-inspired priors.  One
expects albedo and material changes to be correlated, motivating priors such
as piecewise constancy of albedo~\cite{land-mccann:light71,liao:intrinsic13,
barron:intrinsic13,barron:intrinsic15} or sparseness of the set of unique albedo values in a
scene~\cite{omer:color04,gehler:intrinsic11,shen:intrinsic11}.  One also
expects shading to vary smoothly over the image~\cite{garces:intrinsic12}.
Tang~\etal~\cite{hinton:dln12} explore generative learning of priors using
deep belief networks.  Though learning aligns with our philosophy, we take
a discriminative approach.

Systems motivated by physical priors are usually formulated as optimization
routines solving for a point-wise decomposition that satisfies
Equation~\ref{eq:intrin} and also fits with priors imposed over an extended
spatial domain.  Hence, graph-based inference algorithms~\cite{yu:bright09}
and conditional random fields (CRFs) in particular~\cite{IIW} are often used.

\begin{figure*}
   \setlength\fboxsep{0pt}
   \begin{center}
   \begin{minipage}[t]{0.99\linewidth}
      \begin{center}
         \fbox{\includegraphics[width=0.1525\linewidth]{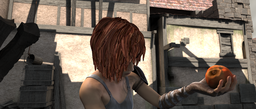}}
         \fbox{\includegraphics[width=0.1525\linewidth]{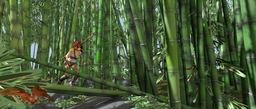}}
         \fbox{\includegraphics[width=0.1525\linewidth]{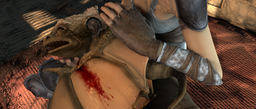}}
         \fbox{\includegraphics[width=0.1525\linewidth]{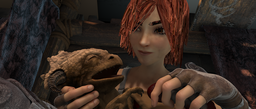}}
         \fbox{\includegraphics[width=0.1525\linewidth]{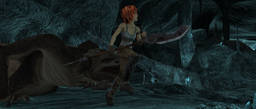}}
         \fbox{\includegraphics[width=0.1525\linewidth]{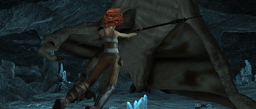}}\\
         \vspace{0.0025\linewidth}
         \fbox{\includegraphics[width=0.1525\linewidth]{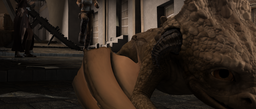}}
         \fbox{\includegraphics[width=0.1525\linewidth]{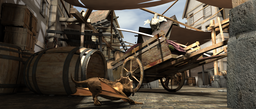}}
         \fbox{\includegraphics[width=0.1525\linewidth]{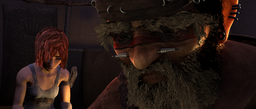}}
         \fbox{\includegraphics[width=0.1525\linewidth]{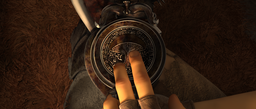}}
         \fbox{\includegraphics[width=0.1525\linewidth]{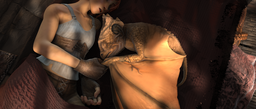}}
         \fbox{\includegraphics[width=0.1525\linewidth]{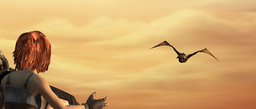}}
      \end{center}
   \end{minipage}
   \end{center}
   \vspace{-0.035\linewidth}
   \begin{center}\rule{0.95\linewidth}{1.0pt}\end{center}
   \vspace{-0.06\linewidth}
   \begin{center}
      \begin{minipage}[t]{0.97\linewidth}
      \begin{center}
      \begin{minipage}[t]{0.19\linewidth}
         \vspace{0pt}
         \begin{center}
            \fbox{\includegraphics[width=1.00\linewidth]{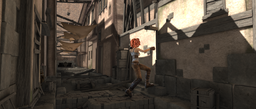}}\\
            \vspace{0.01\linewidth}
            \fbox{\includegraphics[width=1.00\linewidth]{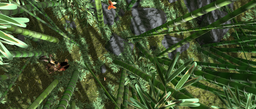}}\\
            \vspace{0.01\linewidth}
            \fbox{\includegraphics[width=1.00\linewidth]{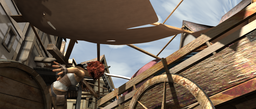}}\\
            \vspace{0.01\linewidth}
            \fbox{\includegraphics[width=1.00\linewidth]{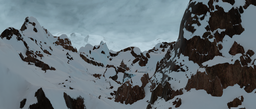}}\\
            \vspace{0.01\linewidth}
            \fbox{\includegraphics[width=1.00\linewidth]{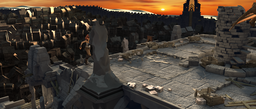}}\\
            \vspace{0.01\linewidth}
            \fbox{\includegraphics[width=1.00\linewidth]{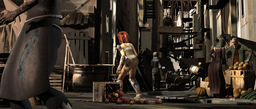}}\\
            \vspace{0.01\linewidth}
            \fbox{\includegraphics[width=1.00\linewidth]{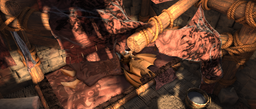}}\\
            \scriptsize{\textbf{\textsf{Image}}}
         \end{center}
      \end{minipage}
      \begin{minipage}[t]{0.19\linewidth}
         \vspace{0pt}
         \begin{center}
            \fbox{\includegraphics[width=1.00\linewidth]{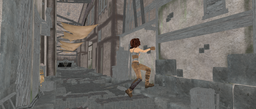}}\\
            \vspace{0.01\linewidth}
            \fbox{\includegraphics[width=1.00\linewidth]{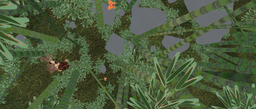}}\\
            \vspace{0.01\linewidth}
            \fbox{\includegraphics[width=1.00\linewidth]{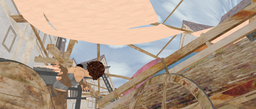}}\\
            \vspace{0.01\linewidth}
            \fbox{\includegraphics[width=1.00\linewidth]{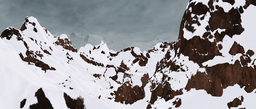}}\\
            \vspace{0.01\linewidth}
            \fbox{\includegraphics[width=1.00\linewidth]{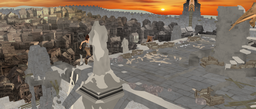}}\\
            \vspace{0.01\linewidth}
            \fbox{\includegraphics[width=1.00\linewidth]{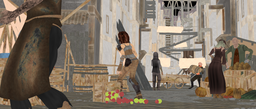}}\\
            \vspace{0.01\linewidth}
            \fbox{\includegraphics[width=1.00\linewidth]{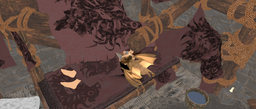}}\\
            \scriptsize{\textbf{\textsf{Ground-truth Albedo}}}
         \end{center}
      \end{minipage}
      \begin{minipage}[t]{0.19\linewidth}
         \vspace{0pt}
         \begin{center}
            \fbox{\includegraphics[width=1.00\linewidth]{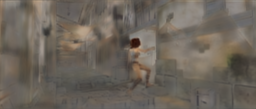}}\\
            \vspace{0.01\linewidth}
            \fbox{\includegraphics[width=1.00\linewidth]{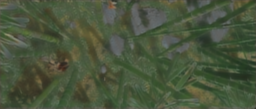}}\\
            \vspace{0.01\linewidth}
            \fbox{\includegraphics[width=1.00\linewidth]{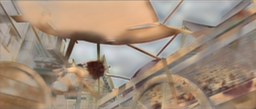}}\\
            \vspace{0.01\linewidth}
            \fbox{\includegraphics[width=1.00\linewidth]{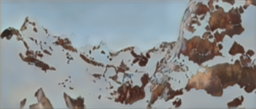}}\\
            \vspace{0.01\linewidth}
            \fbox{\includegraphics[width=1.00\linewidth]{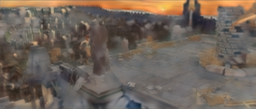}}\\
            \vspace{0.01\linewidth}
            \fbox{\includegraphics[width=1.00\linewidth]{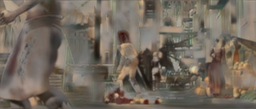}}\\
            \vspace{0.01\linewidth}
            \fbox{\includegraphics[width=1.00\linewidth]{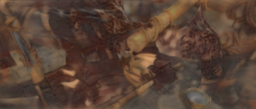}}\\
            \scriptsize{\textbf{\textsf{Our Albedo}}}
         \end{center}
      \end{minipage}
      \begin{minipage}[t]{0.19\linewidth}
         \vspace{0pt}
         \begin{center}
            \fbox{\includegraphics[width=1.00\linewidth]{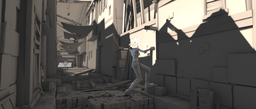}}\\
            \vspace{0.01\linewidth}
            \fbox{\includegraphics[width=1.00\linewidth]{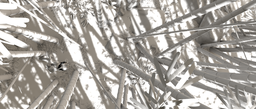}}\\
            \vspace{0.01\linewidth}
            \fbox{\includegraphics[width=1.00\linewidth]{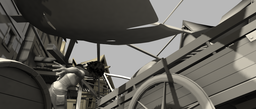}}\\
            \vspace{0.01\linewidth}
            \fbox{\includegraphics[width=1.00\linewidth]{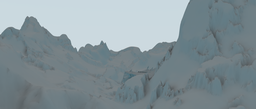}}\\
            \vspace{0.01\linewidth}
            \fbox{\includegraphics[width=1.00\linewidth]{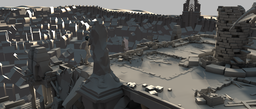}}\\
            \vspace{0.01\linewidth}
            \fbox{\includegraphics[width=1.00\linewidth]{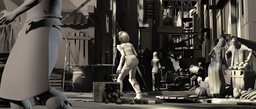}}\\
            \vspace{0.01\linewidth}
            \fbox{\includegraphics[width=1.00\linewidth]{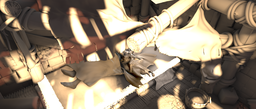}}\\
            \scriptsize{\textbf{\textsf{Ground-truth Shading}}}
         \end{center}
      \end{minipage}
      \begin{minipage}[t]{0.19\linewidth}
         \vspace{0pt}
         \begin{center}
            \fbox{\includegraphics[width=1.00\linewidth]{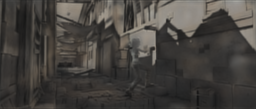}}\\
            \vspace{0.01\linewidth}
            \fbox{\includegraphics[width=1.00\linewidth]{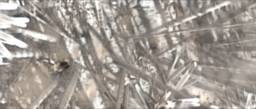}}\\
            \vspace{0.01\linewidth}
            \fbox{\includegraphics[width=1.00\linewidth]{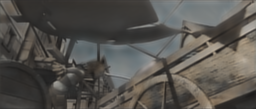}}\\
            \vspace{0.01\linewidth}
            \fbox{\includegraphics[width=1.00\linewidth]{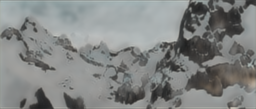}}\\
            \vspace{0.01\linewidth}
            \fbox{\includegraphics[width=1.00\linewidth]{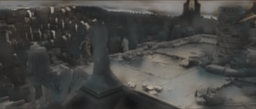}}\\
            \vspace{0.01\linewidth}
            \fbox{\includegraphics[width=1.00\linewidth]{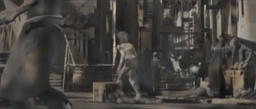}}\\
            \vspace{0.01\linewidth}
            \fbox{\includegraphics[width=1.00\linewidth]{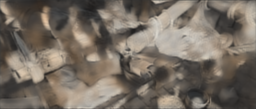}}\\
            \scriptsize{\textbf{\textsf{Our Shading}}}
         \end{center}
      \end{minipage}
      \end{center}
      \end{minipage}
   \end{center}
   \caption{
      \textbf{Albedo-shading decomposition on the MPI Sintel dataset.}
      \emph{Top:}
         A sampling of frames from different scenes comprising the Sintel
         movie.
      \emph{Bottom:}
         Our decomposition results alongside ground-truth albedo and shading
         for some example frames.
   }
   \label{fig:sintel}
\end{figure*}

We forgo both physical modeling constraints and graph-based
inference methods.  Our \emph{direct intrinsics} approach is purely
data-driven and learns a \emph{convolutional regression} which maps a color
image input to its corresponding albedo and shading outputs.  It is
instantiated in the form of a multiscale fully convolutional neural network
(Figure~\ref{fig:overview}).

Key to enabling our direct intrinsics approach is availability of a
large-scale dataset with example ground-truth albedo-shading decompositions.
Unfortunately, collecting such ground-truth for real images is a challenging
task as it requires full control over the lighting environment in which images
are acquired.  This is possible in a laboratory
setting~\cite{grosse:intrinsic09}, but difficult for more realistic scenes.

The Intrinsic Images in the Wild (IIW) dataset~\cite{IIW} attempts to
circumvent the lack of training data through large-scale human labeling
effort.  However, its ground-truth is not in the form of actual decompositions,
but only relative reflectance judgements over a sparse set of point pairs.
These are human judgements rather than physical properties.  They may be
sufficient for training models with strong priors~\cite{IIW}, or most
recently, CNNs for replicating human judgements~\cite{NMY:CVPR:2015}.
But they are insufficient for data-driven learning of intrinsic image
decompositions from scratch.

We circumvent the data availability roadblock by training on purely synthetic
images and testing on both real and synthetic images.  The MPI Sintel
dataset~\cite{Sintel} provides photo-realistic rendered images and
corresponding albedo-shading ground-truth derived from the underlying 3D models
and art assets.  These were first used as training data by Chen and
Koltun~\cite{CK:ICCV:2013} for deriving a more accurate physics-based
intrinsics model.  Figure~\ref{fig:sintel} shows examples.

Section~\ref{sec:direct} describes the details of our CNN architecture and
learning objectives for direct intrinsics.  Our design is motivated by recent
work on using CNNs to recover depth and surface normal estimates from a single
image~\cite{WFG:arXiv:2014,EPF:NIPS:2014,LSLR:arXiv:2015,EF:CVPR:2015}.
Section~\ref{sec:eval} provides experimental results and benchmarks on the
Sintel dataset, and examines the portability of our model to real images.
Section~\ref{sec:conclusion} concludes.

\section{Direct Intrinsics}
\label{sec:direct}

We break the full account of our system into specification of the CNN
architecture, description of the training data, and details of the loss
function during learning.

\begin{figure*}[t]
   \setlength\fboxsep{0pt}
   \begin{center}
      \begin{minipage}[t]{0.48\linewidth}
         \vspace{0pt}
         \begin{center}
            \includegraphics[width=1.0\textwidth]{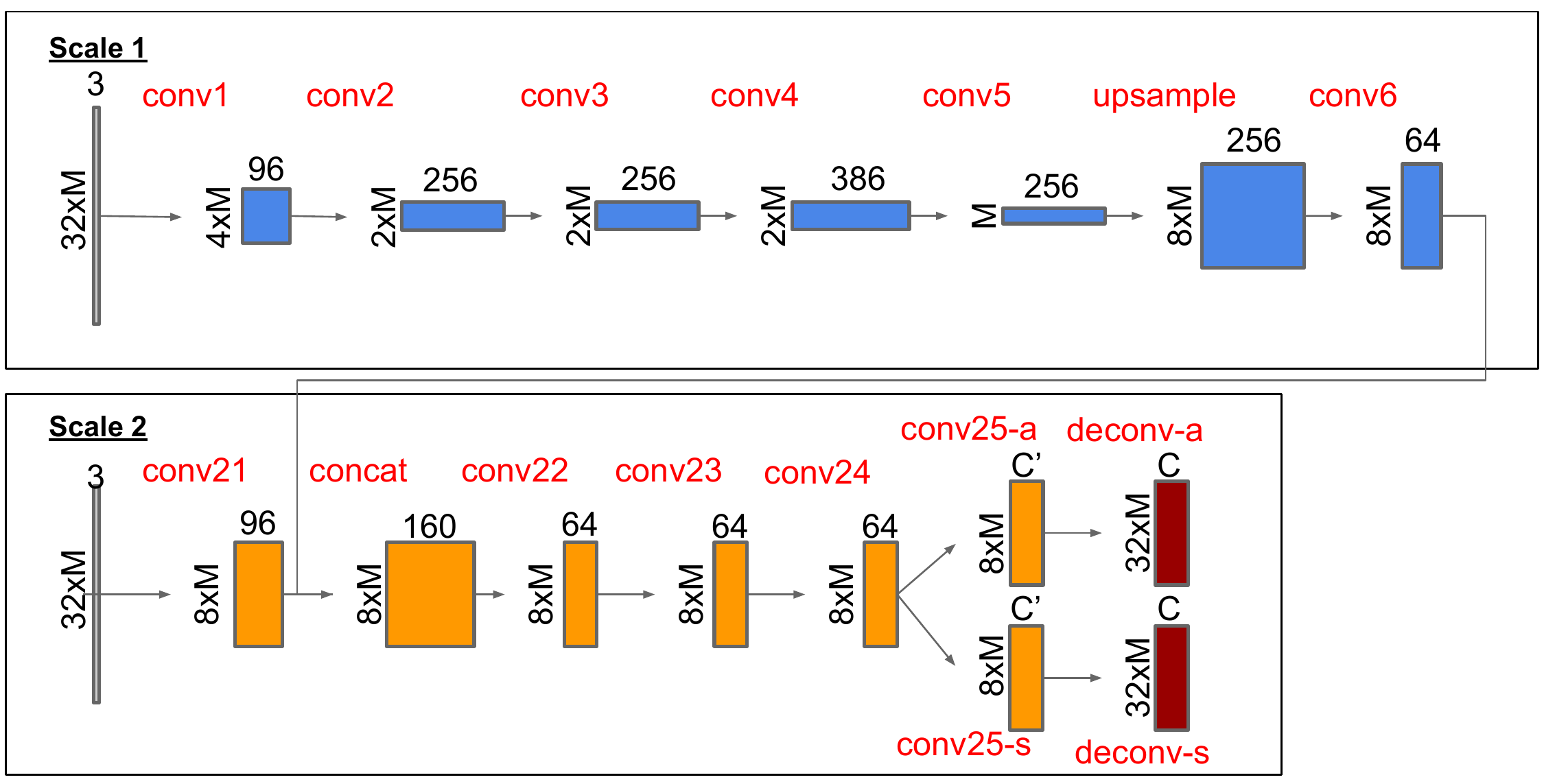}\\
            \scriptsize{\textbf{\textsf{\MABBR}}}
         \end{center}
      \end{minipage}
      \hfill
      \begin{minipage}[t]{0.48\linewidth}
         \vspace{0.07\linewidth}
         \begin{center}
            \begin{overpic}[width=1.0\textwidth,grid=false]{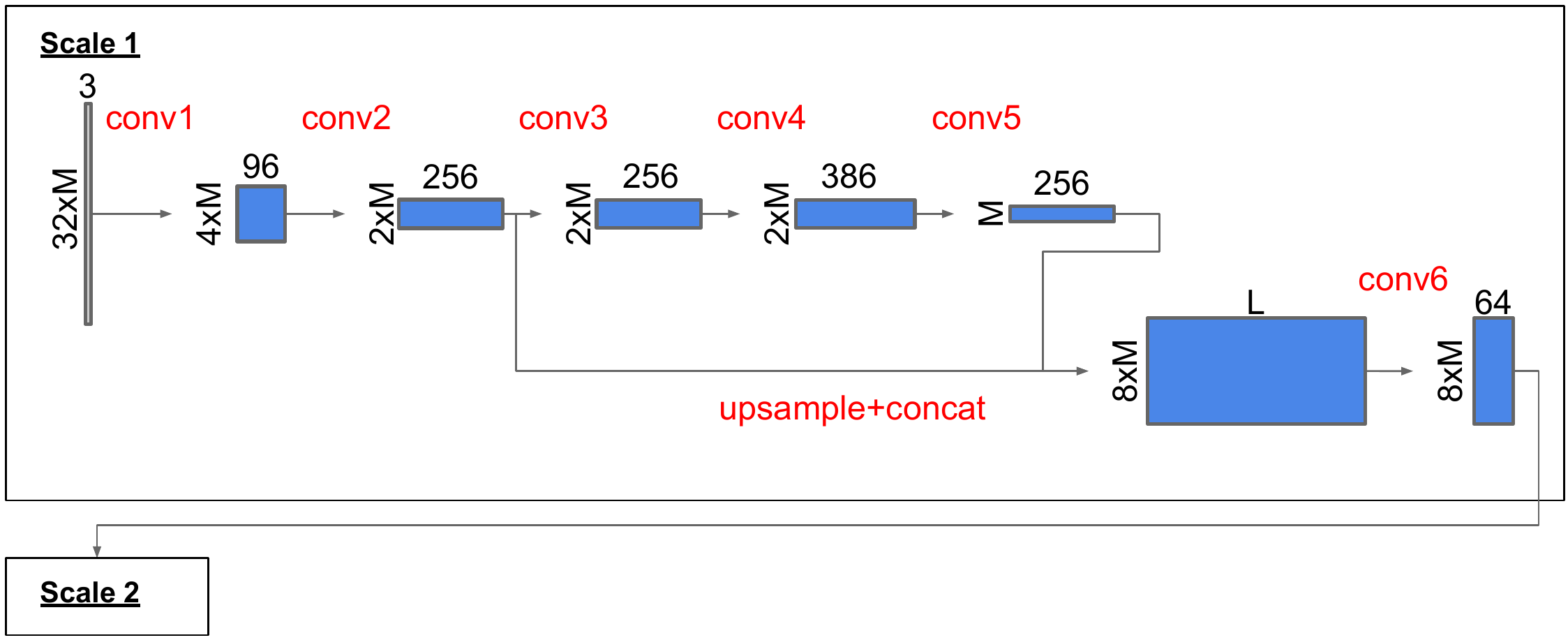}
            \put(78,21.0){\fcolorbox{white}{white}{\textcolor{white}{~-~}}}
            \put(78,21.5){\fcolorbox{white}{white}{\textcolor{white}{~-~}}}
            \end{overpic}
            \scriptsize{\textbf{\textsf{\MABBR+HC}}}
         \end{center}
      \end{minipage}
   \end{center}
   \caption{
      \textbf{CNN architectures.}
         We explore two architectural variants for implementing the direct
         intrinsics network shown in Figure~\ref{fig:overview}.
      \emph{Left:}
         Motivated by the multiscale architecture used by Eigen and
         Fergus~\cite{EF:CVPR:2015} for predicting depth from RGB, we adapt
         a similar network structure for direct prediction of albedo and
         shading from RGB and term it \MNAME~(\MABBR).
      \emph{Right:}
         Recent work~\cite{MYP:ACCV:2014,HAGM:CVPR:2015} shows value in
         directly connecting intermediate layers to the output.  We experiment
         with a version of such connections in the scale~$1$ subnetwork,
         adopting the hypercolumn (HC) terminology~\cite{HAGM:CVPR:2015}.
         The subnetwork for scale~$2$ is identical to that on the left.
         M is input size factor.
   }
   \label{fig:cnns}
\end{figure*}

\subsection{Model}
\label{sec:model}

Intrinsic decomposition requires access to all the precise details of an image
patch as well as overall gist of the entire scene.  The multiscale model of
Eigen and Fergus~\cite{EF:CVPR:2015} for predicting scene depth has these
ingredients and we build upon their network architecture.  In their two-scale
network, they first extract global contextual information in a coarse
subnetwork (scale $1$), and use that subnetwork's output as an
additional input to a finer-scale network (scale $2$).  As
Figure~\ref{fig:cnns} shows, we adopt a {\it \MNAME} (\MABBR) architecture
with important differences from~\cite{EF:CVPR:2015}:

\begin{itemize}
   \item{
      Instead of fully connected layers in scale $1$, we use a $1\times1$
      convolution layer following the upsampling layer.  This choice enables
      our model to run on arbitrary-sized images in a fully convolutional
      fashion.
   }
   \item{
      For nonlinear activations, we use Parametric Rectified Linear Units
      (PReLUs)~\cite{PReLU}.  With PReLUs, a negative slope $a$ for each
      activation map channel appears as a learnable parameter:
      \begin{align}
         g(x_i) &=
            \begin{cases}
               x_i,& x_i \geq 0\\
               a_i x_i,& x_i < 0\\
            \end{cases}
         \label{eq:prelu}
      \end{align}
      where $x_i$ is pre-activation value at $i$-th dimension of a feature map.
      During experiments, we observe better convergence with PReLUs compared
      to ReLUs.
   }
   \item{
      Our network has two outputs, albedo and shading (-a and -s in
      Figure~\ref{fig:cnns}), which it predicts simultaneously.
   }
   \item{
      We optionally use deconvolution to learn to upsample the scale~$2$ output
      to the resolution of the original images~\cite{long_shelhamer_fcn}.
      Without deconvolution, we upsample an RGB output ($C'=C=3$ in
      Figure~\ref{fig:cnns} and the layer between uses fixed bilinear
      interpolation).  With deconvolution, we set $C'=64$ channels,
      $C=3$, and learn to upsample from a richer representation.
   }
\end{itemize}

In addition to these basic changes, we explore a variant of our model, shown
on the right side of Figure~\ref{fig:cnns} that connects multiple layers of
the scale $1$ subnetwork directly to that subnetwork's output.  The reasoning
follows that of Maire~\etal~\cite{MYP:ACCV:2014} and
Hariharan~\etal~\cite{HAGM:CVPR:2015}, with the objective of directly
capturing a representation of the input at multiple levels of abstraction.
We adopt the ''hypercolumn`` (HC) terminology~\cite{HAGM:CVPR:2015} to
designate this modification to \MABBR.

The remaining architectural details are as follows.  For convolutional
layers $1$ through $5$ in the scale $1$ net, we take the common
AlexNet~\cite{alexNet12} design.  Following those, we upsample the feature map 
to a quarter of the original image size, and feed it to a $1\times1$
convolutional layer with $64$-dimensional output (conv6).  Scale $2$ consists
of $4$ convolutional layers for feature extraction followed by albedo and
shading prediction.  The first of these layers has $9\times9$ filters and $96$
output maps.  Subsequently, we concatenate output of the scale $1$ subnetwork
and feed the result into the remaining convolutional and prediction layers, all
of which use $5\times5$ filters.  The optional learned deconvolutional layer
uses $8\times8$ filters with stride $4$.  Whether using deconvolution or
simple upsampling, we evaluate our output on a grid of the same spatial
resolution as the original image.

\subsection{MPI Sintel Dataset}

For training data, we follow Chen and Koltun~\cite{CK:ICCV:2013} and use the
``clean pass'' images of MPI Sintel dataset instead of their ``final'' images,
which are the result of additional computer graphics tricks which distract from
our application.  This eliminates effects such as depth of field, motion blur,
and fog.  Ground-truth shading images are generated by rendering the scene with
all elements assigned a constant grey albedo.

Some images contain defect pixels due to software rendering issues. We
follow~\cite{CK:ICCV:2013} and do not use images with defects in evaluation.
However, limited variation within the Sintel dataset is a concern for
data-driven learning.  Hence, we use defective images in training by
masking out defective pixels (ignoring their contribution to training error). 

\subsection{MIT Intrinsic Image Dataset}

To demonstrate adaptability of our model to the real world images, we use the
MIT intrinsic image dataset~\cite{grosse:intrinsic09}.  Images in this dataset
are acquired with special apparatus, yielding ground-truth reflectance and
shading components for real world objects.  Here, reflectance is synonymous
with our terminology of albedo.

Due to the limited scalability of the collection method, the MIT dataset
contains only $20$ different objects, with each object having $11$ images from
different light sources.  Only $1$ image of $11$ has shading ground-truth.
We generate each of $10$ shading images $S$ from a corresponding original
image $I$ and reflectance image $A$ (identical for all the images because they
are taken from the same object and the same camera settings) by element-wise
division: $S = I' / (\alpha A')$, where $I'$ and $A'$ denote mean values
of RGB channels of $I$ and $A$ respectively, and $\alpha$ is the value that
minimizes the sum of squared error of $I - \alpha A \cdot S$.

For our models trained on MIT, we denote inclusion of these additional
generated examples in training by appending the designation GenMIT to the
model name.  We find that some images in the MIT dataset do not exactly follow
$I=\alpha A \cdot S$, but including these generated shadings still improves
overall performance.

\subsection{Data Synthesis: Matching Sintel to MIT}

Even after generating shading images, the size of the MIT dataset is still
small enough to be problematic for data-driven approaches.  While we can simply
train on Sintel and test on MIT, we observed some differences in dataset
characteristics.  Specifically, the rendering procedure generating Sintel
ground-truth produces output that does not satisfy $I=\alpha A \cdot S$.  In
order to shift the Sintel training data into a domain more representative of
real images, we resynthesized ground-truth $I$ from the ground-truth $A$ and
$S$.  In experiments, we denote this variant by ResynthSintel and find benefit
from training with it when testing on MIT.

\subsection{Data Augmentation}

Throughout all experiments, we crop and mirror training images to generate
additional training examples.  We optionally utilize further data augmentation,
denoted DA in experiments, consisting of scaling and rotating images.

\subsection{Learning}
\label{sec:learning}

Given an image $I$, we denote our dense prediction of albedo $A$ and shading
$S$ maps as:
\begin{equation}
   \begin{aligned}
      (A, S) = F(I, \Theta)
   \end{aligned}
   \label{eq:regression}
\end{equation}
where $\Theta$ consists of all CNN parameters to be learned.

\subsubsection{Scale Invariant L2 Loss}

Since the intensity of our ground-truth albedo and shading is not absolute,
imposing standard regression loss (L2 error) does not work.  Hence, to learn
$\Theta$, we use the scale invariant L2 loss described in~\cite{EF:CVPR:2015}.
Let $Y^\ast$ be a ground-truth image in $\log$ space of either albedo or
shading and $Y$ be a prediction map.  By denoting $y=Y^\ast - Y$ as their
difference, the scale invariant L2 loss is:
\begin{equation}
   \begin{aligned}
      {\cal L}_{\rm SIL2}(Y^\ast, Y) =
         \frac{1}{n}\sum_{i,j,c}y^2_{i,j,c} -
            \lambda \frac{1}{n^2} \left(\sum_{i,j,c}y_{i,j,c}\right)^2
   \end{aligned}
   \label{eq:sil2}
\end{equation}
where $i,j$ are image coordinates, $c$ is the channel index (RGB) and $n$ is
the number of evaluated pixels.  $\lambda$ is a coefficient for balancing the
scale invariant term: it is simply least square loss when $\lambda = 0$, scale
invariant loss when $\lambda = 1$, and an average of the two when $\lambda=0.5$.
We select $\lambda=0.5$ for training on MIT or Sintel separately, as it has
been found to produce good absolute-scale predictions while slightly improving
qualitative output~\cite{EF:CVPR:2015}.  We select $\lambda=1$ for training on
MIT and Sintel jointly, as the intensity scales from the two datasets differ
and the generated images no longer preserve the original intensity scale.
Note that $n$ is not necessarily equal to the number of image pixels because
we ignore defective pixels in the training set.

The loss function for our \MABBR~model is:
\begin{equation}
   \begin{aligned}
      {\cal L}(A^\ast, S^\ast, A, S) =
         {\cal L}_{\rm SIL2}(A^\ast, A) + {\cal L}_{\rm SIL2}(S^\ast, S)
   \end{aligned}
   \label{eq:baselineloss}
\end{equation}

\subsubsection{Gradient L2 Loss}

We also consider training with a loss that favors recovery of piecewise
constant output.  To do so, we use the gradient loss, which is an L2 error
loss between the gradient of prediction and that of the ground-truth.
By letting $\nabla_i$ and $\nabla_j$ be derivative operators in the
$i$- and $j$-dimensions, respectively, of an image, the gradient L2 loss is:
\begin{equation}
   \begin{aligned}
      {\cal L}_{\rm grad}(Y^\ast, Y) =
         \frac{1}{n}\sum_{i,j,c}
            \left[\nabla_i y_{i,j,c} ^2 + \nabla_j y_{i,j,c} ^2\right]
   \end{aligned}
   \label{eq:gradloss}
\end{equation}
Shading cannot be assumed piecewise constant; we do not use gradient loss
for it.  Our objective with gradient loss is:
\begin{equation}
   \begin{aligned}
      {\cal L}(A^\ast&, S^\ast, A, S)=\\
       &{\cal L}_{\rm SIL2}(A^\ast, A)
      + {\cal L}_{\rm SIL2}(S^\ast, S)
      + {\cal L}_{\rm grad}(A^\ast, A)
   \end{aligned}
   \label{eq:gradlossall}
\end{equation}
We denote as \MABBR+GL the version of our model using it.

\subsubsection{Dropout}

Though large compared to other datasets for intrinsic image decomposition,
MPI Sintel, with $890$ examples, is still small compared to the large-scale
datasets for image classification~\cite{deng2009imagenet} on which deep
networks have seen success.  We find it necessary to add additional
regularization during training and employ dropout~\cite{dropout} with
probability $0.5$ for all convolutional layers except conv1 though conv5 in
scale~$1$.

\subsection{Implementation Details}

We implement our algorithms in the Caffe framework~\cite{caffe14}.  We use
stochastic gradient descent with random initialization and momentum of $0.9$
to optimize our networks.  Learning rates for each layer are tuned by hand to
get reasonable convergences.  We train networks with batch size $32$
for $8000$ to $50000$ mini-batch iterations (depending on convergence speed
and dataset).  We randomly crop images at a size of $416 \times 416$ pixels
and mirror them horizontally.  For additional data augmentation (DA), we also
randomly rotate images in the range of [$-15$, $15$] degrees and zoom by a
random factor in the range [$0.8$, $1.2$].

Due to the architecture of our scale $1$ subnetwork, our CNN may take as input
any image whose width and height are each a multiple of $32$ pixels.  For
testing, we pad the images to fit this requirement and then crop the output
map to the original input size. 

\section{Empirical Evaluation}
\label{sec:eval}

\begin{table*}
   \begin{center}
   \begin{footnotesize}
   \begin{tabular}{l||c|c|c||c|c|c||c|c|c}
      \multicolumn{1}{l||}{Sintel Training \& Testing: Image Split}
                                 & \multicolumn{3}{c||}{MSE}                         & \multicolumn{3}{c||}{LMSE}                          & \multicolumn{3}{c}{DSSIM}\\
                                 & Albedo        & Shading         & Avg             & Albedo          & Shading         & Avg             & Albedo          & Shading         & Avg\\
      \hline
      Baseline:
      Shading Constant
                                 & $0.0531$      & $0.0488$        & $0.0510$        & $0.0326$        & $0.0284$        & $0.0305$        & $0.2140$        & $0.2060$        & $0.2100$\\
      Baseline:
      Albedo Constant
                                 & $0.0369$      & $0.0378$        & $0.0374$        & $0.0240$        & $0.0303$        & $0.0272$        & $0.2280$        & $0.1870$        & $0.2075$\\
      Retinex
      \cite{grosse:intrinsic09}  & $0.0606$      & $0.0727$        & $0.0667$        & $0.0366$        & $0.0419$        & $0.0393$        & $0.2270$        & $0.2400$        & $0.2335$\\
      Lee~\etal
      \cite{Lee12}
                                 & $0.0463$      & $0.0507$        & $0.0485$        & $0.0224$        & $0.0192$        & $0.0208$        & $0.1990$        & $0.1770$        & $0.1880$\\
      Barron~\etal
      \cite{barron:intrinsic15}
                                 & $0.0420$      & $0.0436$        & $0.0428$        & $0.0298$        & $0.0264$        & $0.0281$        & $0.2100$        & $0.2060$        & $0.2080$\\
      Chen and Koltun
      \cite{CK:ICCV:2013}
                                 & $0.0307$      & $0.0277$        & $0.0292$        & $0.0185$        & $0.0190$        & $0.0188$        & $\bf{0.1960}$   & $0.1650$        & $0.1805$\\
      \MABBR+dropout+GL          & $\bf{0.0100}$ & $\bf{0.0092}$   & $\bf{0.0096}$   & $\bf{0.0083}$   & $\bf{0.0085}$   & $\bf{0.0084}$   & $0.2014$        & $\bf{0.1505}$   & $\bf{0.1760}$\\
      \multicolumn{10}{c}{\vspace{0.0005\linewidth}}\\
      \multicolumn{1}{l||}{Sintel Training \& Testing: Scene Split}
                                 & \multicolumn{3}{c||}{MSE}                         & \multicolumn{3}{c||}{LMSE}                          & \multicolumn{3}{c}{DSSIM}\\
                                 & Albedo        & Shading         & Avg             & Albedo          & Shading         & Avg             & Albedo          & Shading         & Avg\\
      \hline
      \MABBR                     & $0.0238$      & $0.0250$        & $0.0244$        & $0.0155$        & $0.0172$        & $0.0163$        & $0.2226$        & $0.1816$        & $0.2021$\\
      \MABBR+dropout             & $0.0228$      & $0.0240$        & $0.0234$        & $0.0147$        & $0.0168$        & $0.0158$        & $0.2192$        & $0.1746$        & $0.1969$\\
      \MABBR+dropout+HC          & $0.0231$      & $0.0247$        & $0.0239$        & $0.0147$        & $0.0167$        & $0.0157$        & $0.2187$        & $0.1750$        & $0.1968$\\
      \MABBR+dropout+GL          & $0.0219$      & $0.0242$        & $0.0231$        & $0.0143$        & $0.0166$        & $0.0154$        & $0.2163$        & $0.1737$        & $0.1950$\\
      \MABBR+dropout+deconv+DA   & $0.0209$      & $\bf{0.0221}$   & $0.0215$        & $0.0135$        & $\bf{0.0144}$   & $\bf{0.0139}$   & $0.2081$        & $0.1608$        & $0.1844$\\
      {\hspace{-4.3pt}}$^*$\MABBR+dropout+deconv+DA+GenMIT
                                 & $\bf{0.0201}$ & $0.0224$        & $\bf{0.0213}$   & $\bf{0.0131}$   & $0.0148$        & $\bf{0.0139}$   & $\bf{0.2073}$   & $\bf{0.1594}$   & $\bf{0.1833}$\\
      \multicolumn{10}{c}{\vspace{0.0005\linewidth}}\\
   \end{tabular}\\
   {Key:~~~%
      GL = gradient loss \quad%
      HC = hypercolumns \quad%
      DA = data augmentation (scaling, rotation) \quad%
      GenMIT = add MIT w/generated shading to training%
   }
   \end{footnotesize}
   \end{center}
   \caption{
      \textbf{MPI Sintel benchmarks.}
         We report the standard MSE, LMSE, and DSSIM metrics (lower is better)
         as used in~\cite{CK:ICCV:2013}.  The upper table displays test
         performance for the historical split in which frames from Sintel are
         randomly assigned to train or test sets.  Our method significantly
         outperforms competitors.  The lower table compares our architectural
         variations on a more stringent dataset split which ensures that
         images from a single scene are either all in the training set or all
         in the test set.  Figures~\ref{fig:sintel} and~\ref{fig:comparison}
         display results of our starred method.
   }
   \label{tab:eval}
\end{table*}

\begin{table*}
   \begin{center}
   \begin{footnotesize}
   \begin{tabular}{l||c|c|c||c|c|c}
      \multicolumn{1}{l||}{MIT Training \& Testing: Our Split}
                                 & \multicolumn{3}{c||}{MSE}                         & \multicolumn{3}{c}{LMSE}\\
                                 & Albedo        & Shading         & Avg             & Albedo          & Shading         & Total~\cite{grosse:intrinsic09}\\
      \hline
      {\hspace{-4.3pt}}$^*$Ours:
      \MABBR+dropout+deconv+DA+GenMIT
                                 & $0.0105$      & $\bf{0.0083}$   & $0.0094$        & $0.0296$        & $\bf{0.0163}$   & $0.0234$\\
      \hline
      {\hspace{-4.3pt}}$^*$Ours
      without deconv             & $0.0123$      & $0.0135$        & $0.0129$        & $0.0304$        & $0.0164$        & $0.0249$\\
      Ours
      without DA                 & $0.0107$      & $0.0086$        & $0.0097$        & $0.0300$        & $0.0167$        & $0.0239$\\
      Ours
      without GenMIT             & $0.0106$      & $0.0097$        & $0.0102$        & $0.0302$        & $0.0184$        & $0.0252$\\
      \hline
      Ours
      + Sintel                   & $0.0110$      & $0.0103$        & $0.0107$        & $0.0293$        & $0.0182$        & $0.0243$\\
      {\hspace{-4.3pt}}$^*$Ours
      + ResynthSintel            & $\bf{0.0096}$ & $0.0085$        & $\bf{0.0091}$   & $\bf{0.0267}$   & $0.0172$        & $\bf{0.0224}$\\
      \multicolumn{7}{c}{\vspace{0.0005\linewidth}}\\
      \multicolumn{1}{l||}{MIT Training \& Testing: Barron~\etal's Split}
                                 & \multicolumn{3}{c||}{MSE}                         & \multicolumn{3}{c}{LMSE}\\
                                 & Albedo        & Shading         & Avg             & Albedo          & Shading         & Total~\cite{grosse:intrinsic09}\\
      \hline
      Naive Baseline
      (from~\cite{barron:intrinsic15}, uniform shading)
                                 & $0.0577$      & $0.0455$        & $0.0516$        & $-$             & $-$             & $0.0354$\\
      Barron~\etal
      \cite{barron:intrinsic15}  & $\bf{0.0064}$ & $0.0098$        & $\bf{0.0081}$   & $\bf{0.0162}$   & $\bf{0.0075}$   & $\bf{0.0125}$\\
      Ours
      + ResynthSintel            & $0.0096$      & $\bf{0.0080}$   & $0.0088$        & $0.0275$        & $0.0152$        & $0.0218$\\
      \multicolumn{7}{c}{\vspace{0.0005\linewidth}}\\
   \end{tabular}\\
   {Key:~~~%
      DA = data augmentation (scaling, rotation) \quad%
      GenMIT / Sintel / ResynthSintel = add MIT generated shading / Sintel / resynthesized Sintel to training%
   }
   \end{footnotesize}
   \end{center}
   \caption{
      \textbf{MIT Intrinsic benchmarks.}
         On the real images of the MIT dataset~\cite{grosse:intrinsic09}, our
         system is competitive with Barron~\etal~\cite{barron:intrinsic15}
         according to MSE, but lags behind in LMSE.  Ablated variants (upper
         table, middle rows) highlight the importance of replacing upsampling
         with learned deconvolutional layers.  Variants using additional
         sources of training data (upper table, bottom rows) show gain from
         training with resynthesized Sintel ground-truth that obeys the same
         invariants as the MIT data.  Note that the last column displays the
         reweighted LMSE score according to~\cite{grosse:intrinsic09} rather
         than the simple average.  For visual comparison between results of
         the starred methods, see Figure~\ref{fig:results_mit}.
   }
   \label{tab:eval_mit}
\end{table*}

MPI Sintel dataset:  We use a total of $890$ images in the Sintel
albedo/shading dataset, from $18$ scenes with $50$ frames each
(one of the scenes has only $40$ frames).  We use two-fold cross validation,
that is, training on half of the images and testing on the remaining images, to
obtain our test results on all $890$ images.  Our training/testing split is a
{\it scene split}, placing an entire scene (all images it contains) either
completely in training or completely in testing.  For comparison to prior
work, we retrain with the less stringent historically-used {\it image split}
of Chen and Koltun~\cite{CK:ICCV:2013}, which randomly assigns each image to
the train/test set.

MIT-intrinsic image dataset:  MIT has $20$ objects with $11$ different light
source images, for $220$ images total.  For MIT-intrinsic evaluation, we also 
split into two and use two-fold cross validation.  Following best practices,
we split the validation set by objects rather than images.

We adopt the same three error measures as \cite{CK:ICCV:2013}:
\begin{description}
\item{MSE} is the mean-squared error between albedo/shading results and their
ground-truth.  Following~\cite{grosse:intrinsic09,CK:ICCV:2013}, we use
scale-invariant measures when benchmarking intrinsics results; the
absolute brightness of each image is adjusted to minimize the error. 
\item{LMSE} is the local mean-squared error, which is the average of the
scale-invariant MSE errors computed on overlapping square windows of size
10\% of the image along its larger dimension.
\item{DSSIM} is the dissimilarity version of the {\it structural similarity
index} (SSIM), defined as $\frac{1-\mathrm{SSIM}}{2}$.  SSIM characterizes
image similarity as perceived by human observers.  It combines errors from
independent aspects of luminance, contrast, and structure, which are captured
by mean, variance, and covariance of patches.
\end{description}

On Sintel, we compare our model with two trivial decomposition baselines where
either shading or albedo is assumed uniform grey, the classical Retinex
algorithm (\cite{grosse:intrinsic09} version) which obtains intrinsics by
thresholding gradients, and three state-of-the-art intrinsics approaches which
use not only RGB image input but also depth input.
Barron~\etal~\cite{barron:intrinsic15} estimate the most likely intrinsics
using a shading rendering engine and learned priors on shapes and
illuminations.  Lee~\etal~\cite{Lee12} estimate intrinsic image sequences from
RGB+D video subject to additional shading and temporal constraints.  Chen and
Koltun~\cite{CK:ICCV:2013} use a refined shading model by decomposing it into
direct irradiance, indirect irradiance, and a color component.  On MIT, we
compare with Barron~\etal~\cite{barron:intrinsic15} as well as the trivial
baseline.

\subsection{Results}

\begin{figure*}
   \setlength\fboxsep{0pt}
   \begin{center}
   \begin{minipage}[t]{0.42\linewidth}
      \begin{center}
         \begin{minipage}[t]{0.03\linewidth}
            \vspace{0pt}
            \begin{sideways}\scriptsize{\textbf{\textsf{Input}}}~~~~~~~~~\end{sideways}
         \end{minipage}
         \begin{minipage}[t]{0.47\linewidth}
            \vspace{0pt}
            \begin{center}
               \fbox{\includegraphics[width=1.00\linewidth]{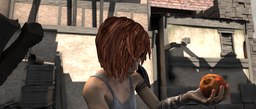}}\\
               \scriptsize{\textbf{\textsf{Color}}}
            \end{center}
         \end{minipage}
         \begin{minipage}[t]{0.47\linewidth}
            \vspace{0pt}
            \begin{center}
               \fbox{\includegraphics[width=1.00\linewidth]{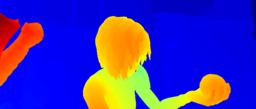}}\\
               \scriptsize{\textbf{\textsf{Depth}}}
            \end{center}
         \end{minipage}
      \end{center}
   \end{minipage}
   \begin{minipage}[t]{0.42\linewidth}
      \begin{center}
         \begin{minipage}[t]{0.47\linewidth}
            \vspace{0pt}
            \begin{center}
               \fbox{\includegraphics[width=1.00\linewidth]{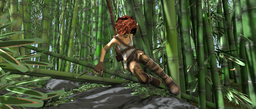}}\\
               \scriptsize{\textbf{\textsf{Color}}}
            \end{center}
         \end{minipage}
         \begin{minipage}[t]{0.47\linewidth}
            \vspace{0pt}
            \begin{center}
               \fbox{\includegraphics[width=1.00\linewidth]{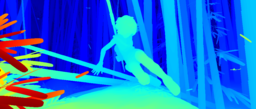}}\\
               \scriptsize{\textbf{\textsf{Depth}}}
            \end{center}
         \end{minipage}
      \end{center}
   \end{minipage}\\
   \vspace{0.01\linewidth}
   \begin{minipage}[t]{0.42\linewidth}
      \begin{center}
         \begin{minipage}[t]{0.03\linewidth}
            \vspace{0pt}
            \begin{sideways}
               \scriptsize{\textbf{\textsf{Ours}}}~~~~~
               \scriptsize{\textbf{\textsf{Chen\hspace{1pt}\&\hspace{1pt}Koltun}}}~
               \scriptsize{\textbf{\textsf{Barron~\etal}}}~~~~
               \scriptsize{\textbf{\textsf{Lee~\etal}}}~~~
               \scriptsize{\textbf{\textsf{Ground-truth}}}
            \end{sideways}
         \end{minipage}
         \begin{minipage}[t]{0.47\linewidth}
            \vspace{0pt}
            \begin{center}
               \fbox{\includegraphics[width=1.00\linewidth]{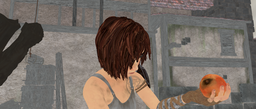}}\\
               \vspace{0.01\linewidth}
               \fbox{\includegraphics[width=1.00\linewidth]{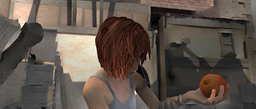}}\\
               \vspace{0.01\linewidth}
               \fbox{\includegraphics[width=1.00\linewidth]{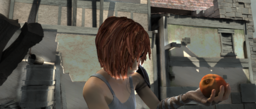}}\\
               \vspace{0.01\linewidth}
               \fbox{\includegraphics[width=1.00\linewidth]{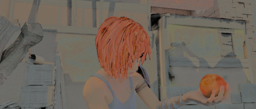}}\\
               \vspace{0.01\linewidth}
               \fbox{\includegraphics[width=1.00\linewidth]{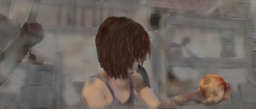}}\\
               \scriptsize{\textbf{\textsf{Albedo}}}
            \end{center}
         \end{minipage}
         \begin{minipage}[t]{0.47\linewidth}
            \vspace{0pt}
            \begin{center}
               \fbox{\includegraphics[width=1.00\linewidth]{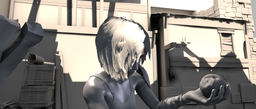}}\\
               \vspace{0.01\linewidth}
               \fbox{\includegraphics[width=1.00\linewidth]{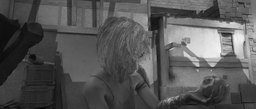}}\\
               \vspace{0.01\linewidth}
               \fbox{\includegraphics[width=1.00\linewidth]{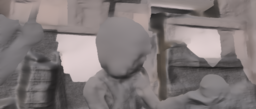}}\\
               \vspace{0.01\linewidth}
               \fbox{\includegraphics[width=1.00\linewidth]{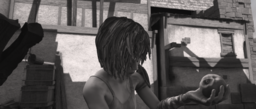}}\\
               \vspace{0.01\linewidth}
               \fbox{\includegraphics[width=1.00\linewidth]{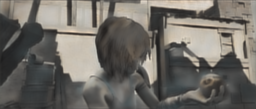}}\\
               \scriptsize{\textbf{\textsf{Shading}}}
            \end{center}
         \end{minipage}
      \end{center}
   \end{minipage}
   \begin{minipage}[t]{0.42\linewidth}
      \begin{center}
         \begin{minipage}[t]{0.47\linewidth}
            \vspace{0pt}
            \begin{center}
               \fbox{\includegraphics[width=1.00\linewidth]{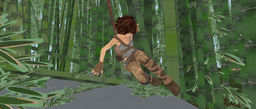}}\\
               \vspace{0.01\linewidth}
               \fbox{\includegraphics[width=1.00\linewidth]{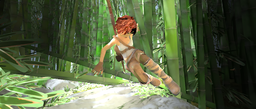}}\\
               \vspace{0.01\linewidth}
               \fbox{\includegraphics[width=1.00\linewidth]{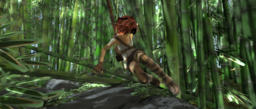}}\\
               \vspace{0.01\linewidth}
               \fbox{\includegraphics[width=1.00\linewidth]{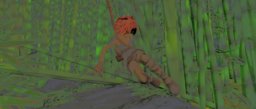}}\\
               \vspace{0.01\linewidth}
               \fbox{\includegraphics[width=1.00\linewidth]{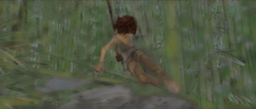}}\\
               \scriptsize{\textbf{\textsf{Albedo}}}
            \end{center}
         \end{minipage}
         \begin{minipage}[t]{0.47\linewidth}
            \vspace{0pt}
            \begin{center}
               \fbox{\includegraphics[width=1.00\linewidth]{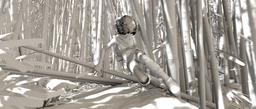}}\\
               \vspace{0.01\linewidth}
               \fbox{\includegraphics[width=1.00\linewidth]{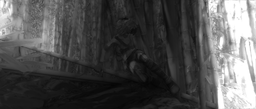}}\\
               \vspace{0.01\linewidth}
               \fbox{\includegraphics[width=1.00\linewidth]{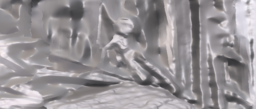}}\\
               \vspace{0.01\linewidth}
               \fbox{\includegraphics[width=1.00\linewidth]{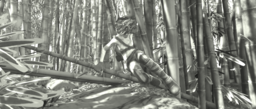}}\\
               \vspace{0.01\linewidth}
               \fbox{\includegraphics[width=1.00\linewidth]{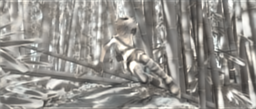}}\\
               \scriptsize{\textbf{\textsf{Shading}}}
            \end{center}
         \end{minipage}
      \end{center}
   \end{minipage}
   \end{center}
   \vspace{-0.025\linewidth}
   \begin{center}\rule{0.83\linewidth}{1.0pt}\end{center}
   \vspace{-0.05\linewidth}
   \begin{center}
   \begin{minipage}[t]{0.42\linewidth}
      \begin{center}
         \begin{minipage}[t]{0.03\linewidth}
            \vspace{0pt}
            \begin{sideways}\scriptsize{\textbf{\textsf{Input}}}~~~~~~~~~\end{sideways}
         \end{minipage}
         \begin{minipage}[t]{0.47\linewidth}
            \vspace{0pt}
            \begin{center}
               \fbox{\includegraphics[width=1.00\linewidth]{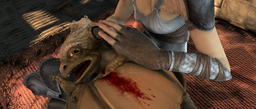}}\\
               \scriptsize{\textbf{\textsf{Color}}}
            \end{center}
         \end{minipage}
         \begin{minipage}[t]{0.47\linewidth}
            \vspace{0pt}
            \begin{center}
               \fbox{\includegraphics[width=1.00\linewidth]{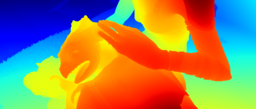}}\\
               \scriptsize{\textbf{\textsf{Depth}}}
            \end{center}
         \end{minipage}
      \end{center}
   \end{minipage}
   \begin{minipage}[t]{0.42\linewidth}
      \begin{center}
         \begin{minipage}[t]{0.47\linewidth}
            \vspace{0pt}
            \begin{center}
               \fbox{\includegraphics[width=1.00\linewidth]{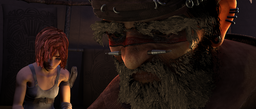}}\\
               \scriptsize{\textbf{\textsf{Color}}}
            \end{center}
         \end{minipage}
         \begin{minipage}[t]{0.47\linewidth}
            \vspace{0pt}
            \begin{center}
               \fbox{\includegraphics[width=1.00\linewidth]{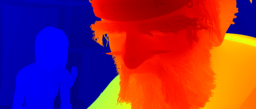}}\\
               \scriptsize{\textbf{\textsf{Depth}}}
            \end{center}
         \end{minipage}
      \end{center}
   \end{minipage}\\
   \vspace{0.01\linewidth}
   \begin{minipage}[t]{0.42\linewidth}
      \begin{center}
         \begin{minipage}[t]{0.03\linewidth}
            \vspace{0pt}
            \begin{sideways}
               \scriptsize{\textbf{\textsf{Ours}}}~~~~~
               \scriptsize{\textbf{\textsf{Chen\hspace{1pt}\&\hspace{1pt}Koltun}}}~
               \scriptsize{\textbf{\textsf{Barron~\etal}}}~~~~
               \scriptsize{\textbf{\textsf{Lee~\etal}}}~~~
               \scriptsize{\textbf{\textsf{Ground-truth}}}
            \end{sideways}
         \end{minipage}
         \begin{minipage}[t]{0.47\linewidth}
            \vspace{0pt}
            \begin{center}
               \fbox{\includegraphics[width=1.00\linewidth]{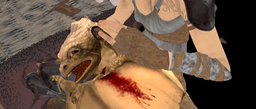}}\\
               \vspace{0.01\linewidth}
               \fbox{\includegraphics[width=1.00\linewidth]{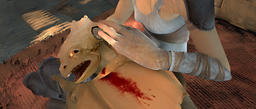}}\\
               \vspace{0.01\linewidth}
               \fbox{\includegraphics[width=1.00\linewidth]{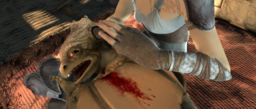}}\\
               \vspace{0.01\linewidth}
               \fbox{\includegraphics[width=1.00\linewidth]{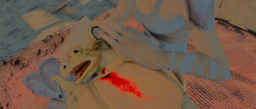}}\\
               \vspace{0.01\linewidth}
               \fbox{\includegraphics[width=1.00\linewidth]{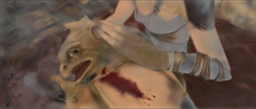}}\\
               \scriptsize{\textbf{\textsf{Albedo}}}
            \end{center}
         \end{minipage}
         \begin{minipage}[t]{0.47\linewidth}
            \vspace{0pt}
            \begin{center}
               \fbox{\includegraphics[width=1.00\linewidth]{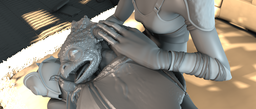}}\\
               \vspace{0.01\linewidth}
               \fbox{\includegraphics[width=1.00\linewidth]{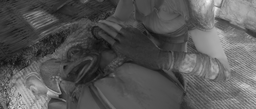}}\\
               \vspace{0.01\linewidth}
               \fbox{\includegraphics[width=1.00\linewidth]{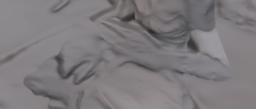}}\\
               \vspace{0.01\linewidth}
               \fbox{\includegraphics[width=1.00\linewidth]{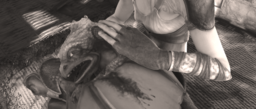}}\\
               \vspace{0.01\linewidth}
               \fbox{\includegraphics[width=1.00\linewidth]{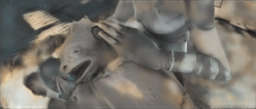}}\\
               \scriptsize{\textbf{\textsf{Shading}}}
            \end{center}
         \end{minipage}
      \end{center}
   \end{minipage}
   \begin{minipage}[t]{0.42\linewidth}
      \begin{center}
         \begin{minipage}[t]{0.47\linewidth}
            \vspace{0pt}
            \begin{center}
               \fbox{\includegraphics[width=1.00\linewidth]{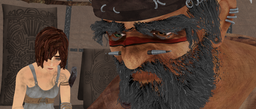}}\\
               \vspace{0.01\linewidth}
               \fbox{\includegraphics[width=1.00\linewidth]{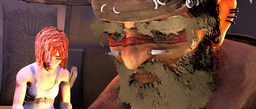}}\\
               \vspace{0.01\linewidth}
               \fbox{\includegraphics[width=1.00\linewidth]{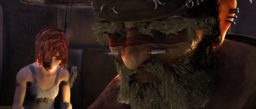}}\\
               \vspace{0.01\linewidth}
               \fbox{\includegraphics[width=1.00\linewidth]{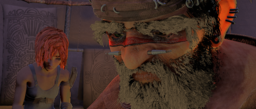}}\\
               \vspace{0.01\linewidth}
               \fbox{\includegraphics[width=1.00\linewidth]{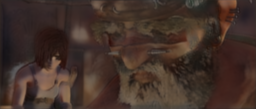}}\\
               \scriptsize{\textbf{\textsf{Albedo}}}
            \end{center}
         \end{minipage}
         \begin{minipage}[t]{0.47\linewidth}
            \vspace{0pt}
            \begin{center}
               \fbox{\includegraphics[width=1.00\linewidth]{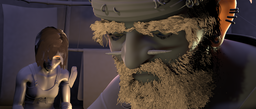}}\\
               \vspace{0.01\linewidth}
               \fbox{\includegraphics[width=1.00\linewidth]{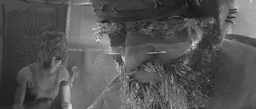}}\\
               \vspace{0.01\linewidth}
               \fbox{\includegraphics[width=1.00\linewidth]{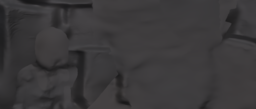}}\\
               \vspace{0.01\linewidth}
               \fbox{\includegraphics[width=1.00\linewidth]{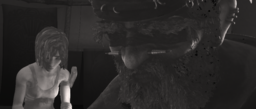}}\\
               \vspace{0.01\linewidth}
               \fbox{\includegraphics[width=1.00\linewidth]{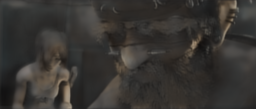}}\\
               \scriptsize{\textbf{\textsf{Shading}}}
            \end{center}
         \end{minipage}
      \end{center}
   \end{minipage}
   \end{center}
   \caption{
      \textbf{Comparison on MPI Sintel.}
      We compare our intrinsic image decompositions with those of
      Lee~\etal~\cite{Lee12},
      Barron~\etal~\cite{barron:intrinsic15}, and
      Chen and Koltun~\cite{CK:ICCV:2013}.
      Our algorithm is unique in \emph{using only RGB and not depth} input
      channels, yet it generates decompositions superior to those of the
      other algorithms, which all rely on full RGB+D input (inverse depth
      shown above).  See Table~\ref{tab:eval} for quantitative benchmarks.
   }
   \label{fig:comparison}
\end{figure*}

\begin{figure*}
   \setlength\fboxsep{0pt}
   \begin{center}
      \begin{minipage}[t]{0.1056\linewidth}
         \vspace{0pt}
         \begin{center}
            \fbox{\includegraphics[width=1.00\linewidth]{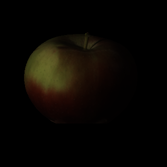}}\\
            \vspace{0.01\linewidth}
            \fbox{\includegraphics[width=1.00\linewidth]{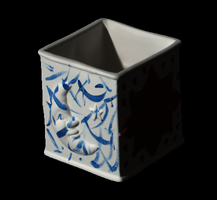}}\\
            \vspace{0.01\linewidth}
            \fbox{\includegraphics[width=1.00\linewidth]{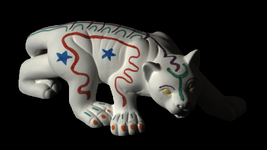}}\\
            \vspace{0.01\linewidth}
            \fbox{\includegraphics[width=1.00\linewidth]{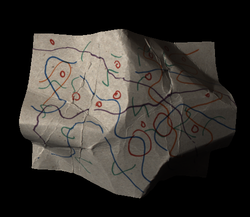}}\\
            \vspace{0.01\linewidth}
            \fbox{\includegraphics[width=1.00\linewidth]{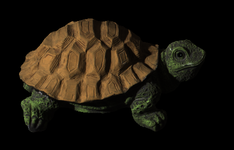}}\\
            \vspace{0.01\linewidth}
            \fbox{\includegraphics[width=1.00\linewidth]{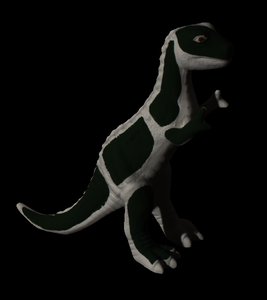}}\\
            \vspace{0.01\linewidth}
            \fbox{\includegraphics[width=1.00\linewidth]{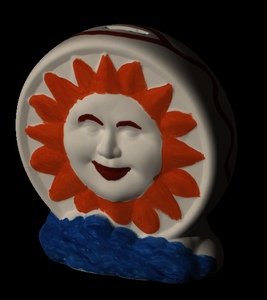}}\\
            \scriptsize{\textbf{\textsf{Image}}}
         \end{center}
      \end{minipage}
      \hfill
      \begin{minipage}[t]{0.44\linewidth}
         \vspace{0pt}
         \begin{center}
         \begin{minipage}[t]{0.24\linewidth}
            \vspace{0pt}
            \begin{center}
               \fbox{\includegraphics[width=1.00\linewidth]{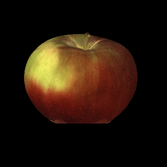}}\\
               \vspace{0.01\linewidth}
               \fbox{\includegraphics[width=1.00\linewidth]{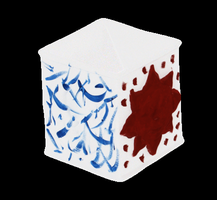}}\\
               \vspace{0.01\linewidth}
               \fbox{\includegraphics[width=1.00\linewidth]{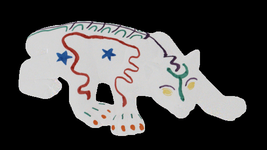}}\\
               \vspace{0.01\linewidth}
               \fbox{\includegraphics[width=1.00\linewidth]{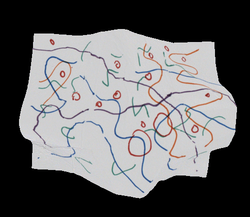}}\\
               \vspace{0.01\linewidth}
               \fbox{\includegraphics[width=1.00\linewidth]{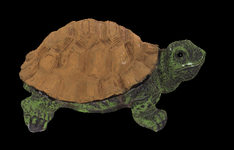}}\\
               \vspace{0.01\linewidth}
               \fbox{\includegraphics[width=1.00\linewidth]{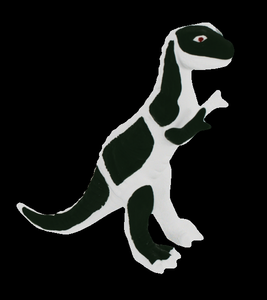}}\\
               \vspace{0.01\linewidth}
               \fbox{\includegraphics[width=1.00\linewidth]{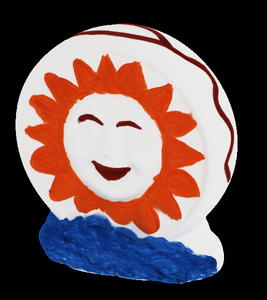}}\\
               \scriptsize{\textbf{\textsf{Ground-truth}}}
            \end{center}
         \end{minipage}
         \begin{minipage}[t]{0.24\linewidth}
            \vspace{0pt}
            \begin{center}
               \fbox{\includegraphics[width=1.00\linewidth]{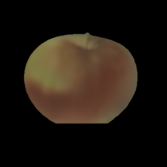}}\\
               \vspace{0.01\linewidth}
               \fbox{\includegraphics[width=1.00\linewidth]{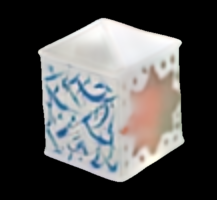}}\\
               \vspace{0.01\linewidth}
               \fbox{\includegraphics[width=1.00\linewidth]{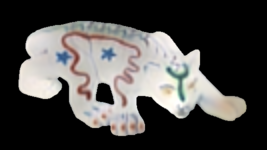}}\\
               \vspace{0.01\linewidth}
               \fbox{\includegraphics[width=1.00\linewidth]{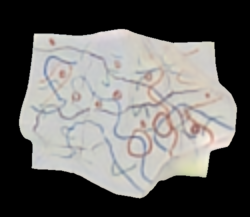}}\\
               \vspace{0.01\linewidth}
               \fbox{\includegraphics[width=1.00\linewidth]{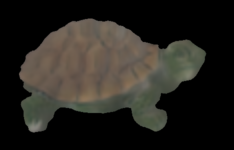}}\\
               \vspace{0.01\linewidth}
               \fbox{\includegraphics[width=1.00\linewidth]{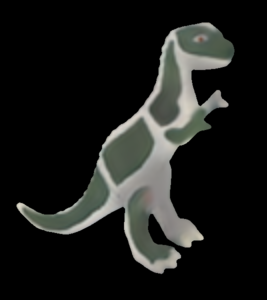}}\\
               \vspace{0.01\linewidth}
               \fbox{\includegraphics[width=1.00\linewidth]{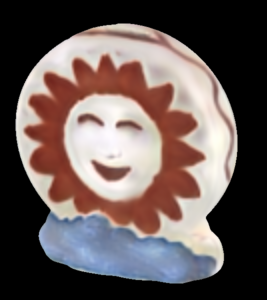}}\\
               \scriptsize{\textbf{\textsf{w/o deconv}}}
            \end{center}
         \end{minipage}
         \begin{minipage}[t]{0.24\linewidth}
            \vspace{0pt}
            \begin{center}
               \fbox{\includegraphics[width=1.00\linewidth]{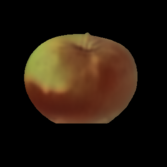}}\\
               \vspace{0.01\linewidth}
               \fbox{\includegraphics[width=1.00\linewidth]{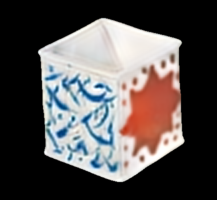}}\\
               \vspace{0.01\linewidth}
               \fbox{\includegraphics[width=1.00\linewidth]{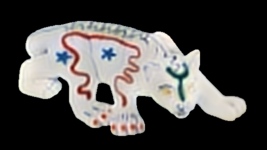}}\\
               \vspace{0.01\linewidth}
               \fbox{\includegraphics[width=1.00\linewidth]{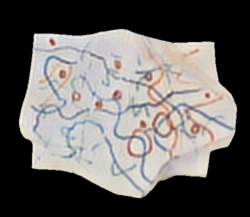}}\\
               \vspace{0.01\linewidth}
               \fbox{\includegraphics[width=1.00\linewidth]{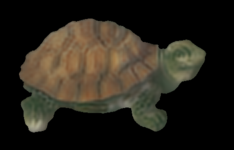}}\\
               \vspace{0.01\linewidth}
               \fbox{\includegraphics[width=1.00\linewidth]{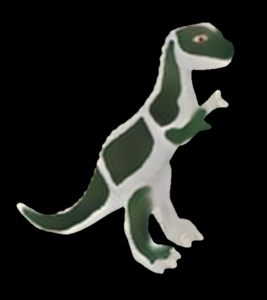}}\\
               \vspace{0.01\linewidth}
               \fbox{\includegraphics[width=1.00\linewidth]{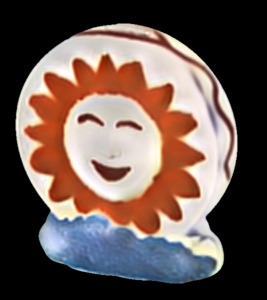}}\\
               \scriptsize{\textbf{\textsf{Base System}}}
            \end{center}
         \end{minipage}
         \begin{minipage}[t]{0.24\linewidth}
            \vspace{0pt}
            \begin{center}
               \fbox{\includegraphics[width=1.00\linewidth]{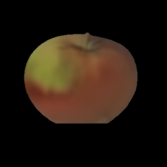}}\\
               \vspace{0.01\linewidth}
               \fbox{\includegraphics[width=1.00\linewidth]{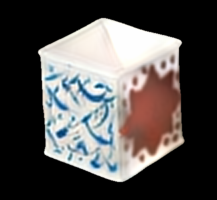}}\\
               \vspace{0.01\linewidth}
               \fbox{\includegraphics[width=1.00\linewidth]{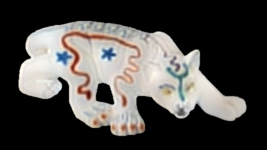}}\\
               \vspace{0.01\linewidth}
               \fbox{\includegraphics[width=1.00\linewidth]{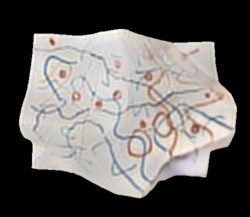}}\\
               \vspace{0.01\linewidth}
               \fbox{\includegraphics[width=1.00\linewidth]{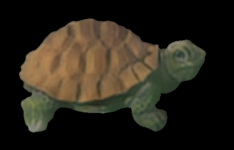}}\\
               \vspace{0.01\linewidth}
               \fbox{\includegraphics[width=1.00\linewidth]{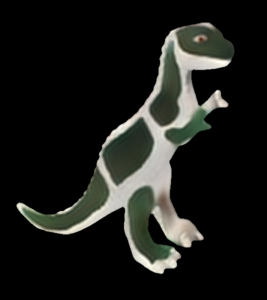}}\\
               \vspace{0.01\linewidth}
               \fbox{\includegraphics[width=1.00\linewidth]{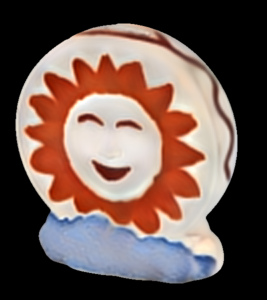}}\\
               \scriptsize{\textbf{\textsf{w/ResynthSintel}}}
            \end{center}
         \end{minipage}\\
         \vspace{5pt}
         \scriptsize{\textbf{\textsf{Albedo}}}
         \end{center}
      \end{minipage}
      \hfill
      \begin{minipage}[t]{0.44\linewidth}
         \vspace{0pt}
         \begin{center}
         \begin{minipage}[t]{0.24\linewidth}
            \vspace{0pt}
            \begin{center}
               \fbox{\includegraphics[width=1.00\linewidth]{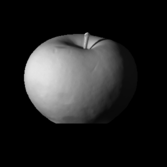}}\\
               \vspace{0.01\linewidth}
               \fbox{\includegraphics[width=1.00\linewidth]{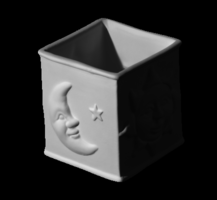}}\\
               \vspace{0.01\linewidth}
               \fbox{\includegraphics[width=1.00\linewidth]{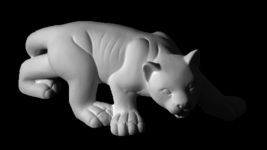}}\\
               \vspace{0.01\linewidth}
               \fbox{\includegraphics[width=1.00\linewidth]{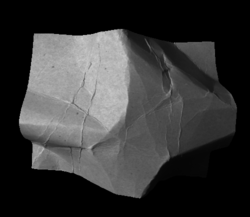}}\\
               \vspace{0.01\linewidth}
               \fbox{\includegraphics[width=1.00\linewidth]{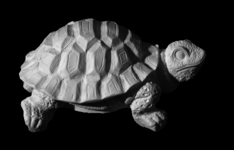}}\\
               \vspace{0.01\linewidth}
               \fbox{\includegraphics[width=1.00\linewidth]{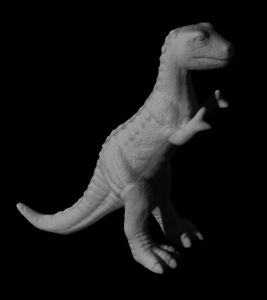}}\\
               \vspace{0.01\linewidth}
               \fbox{\includegraphics[width=1.00\linewidth]{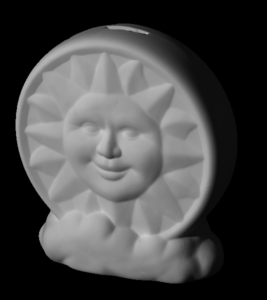}}\\
               \scriptsize{\textbf{\textsf{Ground-truth}}}
            \end{center}
         \end{minipage}
         \begin{minipage}[t]{0.24\linewidth}
            \vspace{0pt}
            \begin{center}
               \fbox{\includegraphics[width=1.00\linewidth]{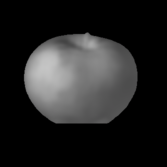}}\\
               \vspace{0.01\linewidth}
               \fbox{\includegraphics[width=1.00\linewidth]{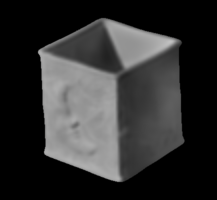}}\\
               \vspace{0.01\linewidth}
               \fbox{\includegraphics[width=1.00\linewidth]{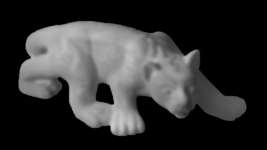}}\\
               \vspace{0.01\linewidth}
               \fbox{\includegraphics[width=1.00\linewidth]{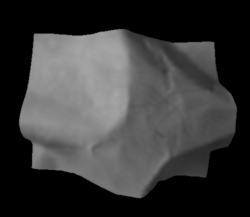}}\\
               \vspace{0.01\linewidth}
               \fbox{\includegraphics[width=1.00\linewidth]{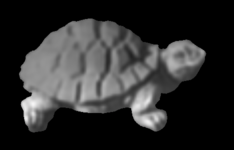}}\\
               \vspace{0.01\linewidth}
               \fbox{\includegraphics[width=1.00\linewidth]{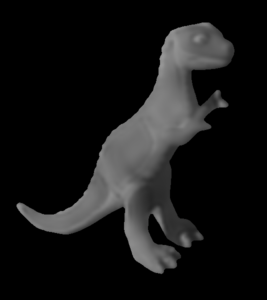}}\\
               \vspace{0.01\linewidth}
               \fbox{\includegraphics[width=1.00\linewidth]{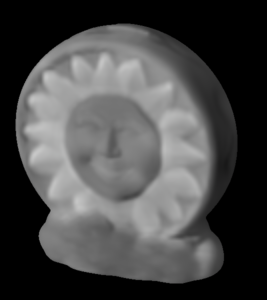}}\\
               \scriptsize{\textbf{\textsf{w/o deconv}}}
            \end{center}
         \end{minipage}
         \begin{minipage}[t]{0.24\linewidth}
            \vspace{0pt}
            \begin{center}
               \fbox{\includegraphics[width=1.00\linewidth]{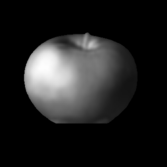}}\\
               \vspace{0.01\linewidth}
               \fbox{\includegraphics[width=1.00\linewidth]{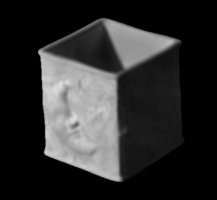}}\\
               \vspace{0.01\linewidth}
               \fbox{\includegraphics[width=1.00\linewidth]{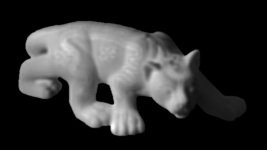}}\\
               \vspace{0.01\linewidth}
               \fbox{\includegraphics[width=1.00\linewidth]{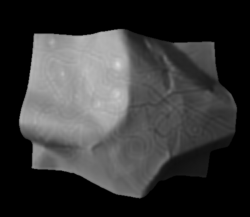}}\\
               \vspace{0.01\linewidth}
               \fbox{\includegraphics[width=1.00\linewidth]{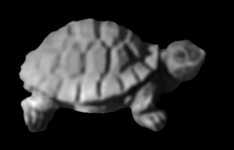}}\\
               \vspace{0.01\linewidth}
               \fbox{\includegraphics[width=1.00\linewidth]{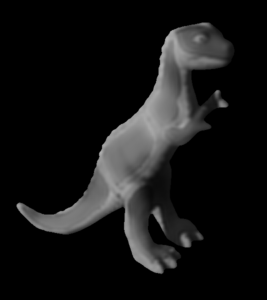}}\\
               \vspace{0.01\linewidth}
               \fbox{\includegraphics[width=1.00\linewidth]{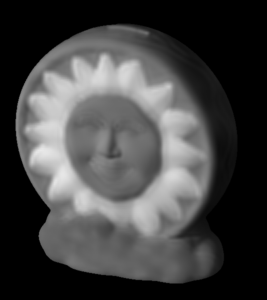}}\\
               \scriptsize{\textbf{\textsf{Base System}}}
            \end{center}
         \end{minipage}
         \begin{minipage}[t]{0.24\linewidth}
            \vspace{0pt}
            \begin{center}
               \fbox{\includegraphics[width=1.00\linewidth]{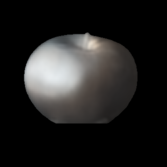}}\\
               \vspace{0.01\linewidth}
               \fbox{\includegraphics[width=1.00\linewidth]{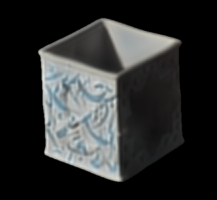}}\\
               \vspace{0.01\linewidth}
               \fbox{\includegraphics[width=1.00\linewidth]{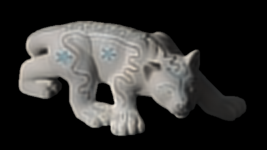}}\\
               \vspace{0.01\linewidth}
               \fbox{\includegraphics[width=1.00\linewidth]{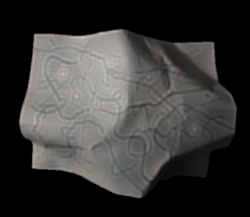}}\\
               \vspace{0.01\linewidth}
               \fbox{\includegraphics[width=1.00\linewidth]{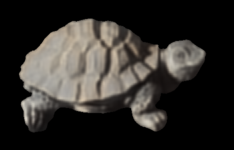}}\\
               \vspace{0.01\linewidth}
               \fbox{\includegraphics[width=1.00\linewidth]{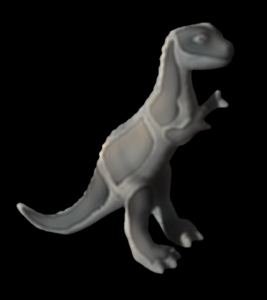}}\\
               \vspace{0.01\linewidth}
               \fbox{\includegraphics[width=1.00\linewidth]{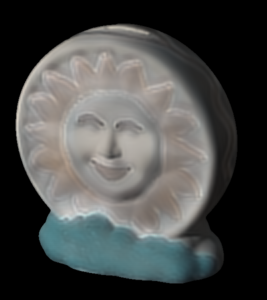}}\\
               \scriptsize{\textbf{\textsf{w/ResynthSintel}}}
            \end{center}
         \end{minipage}\\
         \vspace{5pt}
         \scriptsize{\textbf{\textsf{Shading}}}
         \end{center}
      \end{minipage}
   \end{center}
   \caption{
      \textbf{Adaptability to real images.}
         Our base system (\MABBR+dropout+deconv+DA+GenMIT) produces quite
         reasonable results when trained on the MIT intrinsic image dataset.
         Without learned deconvolution, both albedo and shading quality
         suffer noticeably.  Including resynthesized Sintel data during training
         improves albedo prediction, but biases shading towards Sintel-specific
         lighting conditions.
   }
   \label{fig:results_mit}
\end{figure*}

The top panel of Table~\ref{tab:eval} (image split case) shows that evaluated
on Chen and Koltun's test set, our \MABBR+dropout+GL model significantly
outperforms all competing methods according to MSE and LMSE.  It is also
overall better according to DSSIM than the current state-of-art method of
Chen and Koltun: while our albedo DSSIM is $0.0054$ larger, our shading DSSIM
is $0.0145$ smaller.  Note that Chen and Koltun's method utilizes 
depth information and is also trained on the DSSIM measure directly,
whereas ours is based on the color image alone and is not trained to optimize
the DSSIM score.

The bottom panel of Table~\ref{tab:eval} (scene-split case) is more indicative
of an algorithm's out-of-sample generalization performance; the test scenes
have not been seen during training.  These results show that: 1) The out-of-sample
errors in the scene-split case are generally larger than the in-sample errors
in the image-split case; 2) While HC has negligible effect, each tweak with
dropout, gradient loss, learned deconvolutional layers, and data augmentation
improves performance; 3) Training on Sintel and MIT together provides a small
improvement when testing on Sintel.

Figure~\ref{fig:sintel} shows sample results from our best model, while 
Figure~\ref{fig:comparison} displays a side-by-side comparison with three
other approaches.  An important distinction is that our results are based on
RGB alone, while the other approaches require both RGB and depth input.  Across
a diversity of scenes, any of the three RGB+D approaches could break down in one
of the scenes on either albedo or shading: \eg Lee~\etal's method on the
bamboo scene, Barron~\etal's method on the dragon scene, Chen and Koltun's
method on the old man scene.  The quality of our results is even across scenes
and remains overall consistent with both albedo and shading ground-truth.

Table~\ref{tab:eval_mit} shows that our model graciously adapts to real images.
Trained on MIT alone, it produces reasonable results.  Naively adding Sintel
data to training hurts performance, but mixing our resynthesized version of
Sintel into training results in noticeable improvements to albedo estimation
when testing on MIT.  The behavior of ablated system variants on MIT mirrors
our findings on Sintel.  On MIT, the learned deconvolutional layer is
especially important.  Output in Figure~\ref{fig:results_mit} exhibits clear
visual degradation upon its removal.  Figure~\ref{fig:results_mit} illustrates
a tradeoff when using resynthesized Sintel training data: there is an overall
benefit, but a Sintel-specific shading prior (bluish tint) leaks in.

In addition to Sintel and MIT, we briefly experiment with testing, but not
training, our models on the IIW dataset~\cite{IIW}.  Here, performance is less
than satisfactory (WHDR=$27.2$), compared to both our own prior
work~\cite{NMY:CVPR:2015} and the current
state-of-the-art~\cite{ZKE:ICCV:2015}, which are trained specifically for IIW.
We speculate that there could be some discrepancy between the tasks of
predicting human reflectance judgements (WHDR metric) and physically-correct
albedo-shading decompositions.  As we observed when moving from Sintel to MIT,
there could be a domain shift between Sintel/MIT and IIW for which we are not
compensating.  We leave these interesting issues for future work.

\section{Conclusion}
\label{sec:conclusion}

We propose {\it direct intrinsics}, a new intrinsic image decomposition
approach that is not based on the physics of image formation or the statistics
of shading and albedo priors, but learns the dual associations between the
image and the albedo+shading components directly from training data.

We develop a two-level feed-forward CNN architecture based on a successful
previous model for RGB to depth prediction, where the coarse level architecture
predicts the global context and the finer network uses the output of the
coarse network to predict the finer resolution result.  Combined with
well-designed loss functions, data augmentation, dropout, and deconvolution,
we demonstrate that direct intrinsics outperforms state-of-the-art methods
that rely not only on more complex priors and graph-based inference, but also
on the additional input of scene depth.

Our data-driven learning approach is more flexible, generalizable, and easier
to model.  It only needs training data, requires no hand-designed features or
representations, and can adapt to unrealistic illuminations and complex
albedo, shape, and lighting patterns.  Our model works with both synthetic and
real images and can further improve on real images when augmenting training
with synthetic examples.

~\\
\noindent
{\small\textbf{Acknowledgments.}
We thank Ayan Chakrabarti for valuable discussion.  We thank both Qifeng Chen
and Jon Barron for providing their code and accompanying support.}

{\small
\bibliographystyle{ieee}
\bibliography{direct_ii}
}

\end{document}